\documentclass[runningheads]{llncs}

\usepackage[T1]{fontenc}
\usepackage{graphicx}
\usepackage{hyperref}
\usepackage{xcolor}
\usepackage{booktabs}
\usepackage{multirow}
\usepackage{multicol}
\usepackage{makecell}
\usepackage{tabularx} 
\usepackage{soul}
\usepackage{colortbl}
\usepackage{algorithm}
\usepackage{algorithmic}

\begin{document}

\title{Practical Insights into Semi-Supervised Object Detection Approaches}

\titlerunning{Practical Insights into Semi-Supervised Object Detection Approaches}

\author{Chaoxin Wang\inst{1,3,\textsuperscript{\textdagger}}
\and
Bharaneeshwar Balasubramaniyam \inst{2}
\and
Anurag Sangem \inst{3}
\and
Nicolais Guevara\inst{3}
\and
Doina Caragea\inst{2}
}

\authorrunning{C. Wang et al.}

\institute{
Department of Computer Science, Dominican University, River Forest, IL 60305\\
\email{\{cwang\}@dom.edu}
\and
Department of Computer Science, Kansas State University, Manhattan, KS 66506\\
\email{\{bharanibala,dcaragea\}@ksu.edu}
\and
Peak Technologies, Littleton, MA 01460\\
\email{\{anurag.sangem,nicolais.guevara\}@peaktech.com}
}

\maketitle

\begin{abstract}
Learning in data-scarce settings has recently gained significant attention in the research community. Semi-supervised object detection (SSOD) aims to improve detection performance by leveraging a large number of unlabeled images alongside a limited number of labeled images (a.k.a., few-shot learning). In this paper, we present a comprehensive comparison of three state-of-the-art SSOD approaches, including MixPL, Semi-DETR and Consistent-Teacher, with the goal of understanding how performance varies with the number of labeled images. We conduct experiments using the MS-COCO and Pascal VOC datasets, two popular object detection benchmarks which allow for standardized evaluation. In addition, we evaluate the SSOD approaches on a custom Beetle dataset which enables us to gain insights into their performance on specialized datasets with a smaller number of object categories. Our findings highlight the trade-offs between accuracy, model size, and latency, providing insights into which methods are best suited for low-data regimes.


\end{abstract}

\begingroup
  \renewcommand\thefootnote{}
  \footnotetext{\textsuperscript{\textdagger} This work was done while the author was at Peak Technologies.}
\endgroup

\section{Introduction}

In many contemporary real-world applications, such as manufacturing, supply chain, agriculture or robotics, object detectors (e.g., YOLOv13 \cite{lei2025yolov13}, Faster R-CNN \cite{Shaoqing_faster_rcnn} or DETR \cite{detr_nicolas}) must be used to recognize dozens or even hundreds of categories of objects in complex scenes. Accurate object detectors rely heavily on annotated image datasets \cite{shehzadi2024semi,chen_od_review_2024}. Yet fully annotating images with high-quality bounding boxes for training accurate detectors is prohibitively expensive, requiring significant human effort and tens of thousands of dollars for relatively small datasets \cite{shehzadi2024semi}. For large-scale vision datasets such as MS-COCO \cite{lin_microsoft_2015}, manual labeling is not only costly but also prone to human errors and noise \cite{northcutt_confident_2022}. Moreover, supervised learning models trained from large amounts of labeled data tend to overfit to training distributions, weakening their generalization capabilities, especially when faced with distribution shifts or novel classes \cite{chen_ssuvl_review_2024}.

SSOD addresses these challenges by leveraging large sets of easily-collected, unlabeled images alongside a small set of annotated images, dramatically cutting annotation cost while still learning robust models \cite{wang_ssod_review_2023,shehzadi2024semi}. SSOD methods typically rely on an iterative process which includes: 1) generating pseudo-labels from a teacher model, 2) retraining a student model on a mix of labeled and pseudo-labeled data, 3) repeating these steps over several rounds. In this study, we evaluate three recent state-of-the-art SSOD approaches with publicly available implementations \cite{shehzadi2025step,shehzadi2024semi}: MixPL \cite{chen_mixed_2023}, Semi-DETR \cite{zhang_semi-detr_2023}, and Consistent-Teacher \cite{consistent_teacher}. {We note that while Sparse Semi-DETR \cite{shehzadi2024sparse} and STEP-DETR \cite{shehzadi2025step} represent promising recent directions, they were excluded from this analysis as their implementations were not publicly available at the onset of our study.} While prior work often reports performance using a certain percentage (e.g., 1\% or 10\%) of MS-COCO dataset, the specific number of labeled images or object instances needed for satisfactory performance remains unclear. Therefore, we investigate how these methods perform under explicitly defined, limited numbers of labeled images or object instances per category.

Some recent studies, such as SoftER Teacher \cite{xiong2021semi} and APLDet \cite{tang_apldet_2024} have explored SSOD in the context of Few-Shot Object Detection (FSOD). However, the focus of these works is on identifying novel classes with a few annotated images, and they follow a Two-stage Fine-tuning Approach (TFA) \cite{wang_2020_frustratingly}, where a detector is first trained on base classes for which annotated images are abundant, and subsequently fine-tuned on novel classes with a small number of annotated images. As opposed to aiming to identify novel classes as in FSOD, our focus is on training SSOD detectors to identify all classes pertaining to a task in a low-data regime, with a small number of annotated images per each category. 

We should also note that most of the prior works in SSOD/FSOD focus on performance, but do not report key information such as model size and inference time, which are especially important in real-world applications where resources are generally limited and inference speed is critical. To account for this limitation, in this work, in addition to studying state-of-the-art SSOD approaches in a low-data regime to gain insights into the number of images per category needed for satisfactory performance, we also aim to understand the trade-off between performance and inference time and computation requirements.

More specifically, we investigate the aforementioned methods, MixPL \cite{chen_mixed_2023}, Semi-DETR \cite{zhang_semi-detr_2023}, and Consistent-Teacher \cite{consistent_teacher}, using the popular MS-COCO \cite{lin_microsoft_2015} and Pascal VOC \cite{everingham2015pascal} datasets, which contain images with a variety of common (e.g., person, car, truck) objects in 80 and 20 categories, respectively, as well as a custom Beetle dataset \cite{wang_detecting_2023}, consisting of images with objects in one of 7 beetle categories. Some images in the MS-COCO/Pascal VOC datasets include a diverse set of object categories that co-occur together, and some categories appear in a large number of images (e.g., person). As opposed to that, images in the Beetle dataset contain one or more beetles of the same type (in the same category). We experiment with SSOD models trained from a specific number $k$ of labeled images selected for each category. This ensures that each category appears in at least $k$ images. However, due to the fact that the MS-COCO and Pascal VOC datasets contain objects that co-occur together, some categories appear in more than $k$ images.

Our goal is to gain insights into how performance varies with the number of labeled images. The lessons learned can lead to recommendations regarding the number of images needed to achieve the desired performance on a new dataset.

In addition to the focus on high-performing, effective SSOD models trained with a small number of annotated images, we also aim to identify models that provide a good balance between performance, robustness, as well as inference time/latency. To the best of our knowledge, this is one of the first empirical studies to analyze the behavior of leading SSOD models in this challenging regime. Our goal is to provide insights into their limitations and capabilities when labeled data is extremely limited.

To summarize, we seek to answer the following key research questions: 

\begin{itemize}
  \item RQ1: What is the best SSOD when the number $k$ of labeled images selected for each category varies between 1 and 150? 
  \item RQ2: What are the trade-offs between low-regime data training and overall object detection performance?
  \item RQ3: What are the trade-offs between performance, model size and latency?
\end{itemize}

\section{Background and Related Work}

\subsection{Semi-Supervised Object Detection}

SSOD methods such as MixPL \cite{chen_mixed_2023}, Semi-DETR \cite{zhang_semi-detr_2023} and Consistent-Teacher \cite{consistent_teacher} typically incorporate self-training, pseudo-labeling, consistency regularization or a combination of these techniques along with data augmentation strategies. In most of the semi-supervised tasks, the student model is trained and used for the inference \cite{shehzadi2024semi}. Unlike prior works \cite{chen_mixed_2023,zhang_semi-detr_2023,consistent_teacher} that report results based on dataset percentages (e.g., 1\% or 10\% of COCO), our study standardizes evaluation by fixing the number of labeled images per class in a few-shot manner.

\subsection{Few-Shot Object Detection}

FSOD approaches are categorized into four main paradigms: (1) Data Augmentation, (2) Transfer Learning, (3) Distance Metric Learning, and (4) Meta-Learning \cite{Simone_fewshotsurvey_2022}. Our work intersects with this taxonomy by evaluating SSOD methods under a fixed few-shot regime, although we do not build directly on FSOD-specific approaches. Most FSOD methods rely on meta-learning or transfer learning, and assume novel-class detection \cite{chudasama_fsod_survey_2024}, unlike our setting which aims to detect all categories pertaining to a task of interest in a low-data regime, closely aligned with many practical applications. Leading FSOD models include CD-ViTO \cite{cdvito}, hANMCL \cite{hanmcl_park2022}, UniFS \cite{jin2024unifs}, De-ViT \cite{devit_ZhangLWB24}, and DETReg \cite{detreg_bar2022}.

\subsection{Semi-Supervised Few-Shot Object Detection} 

While recent efforts, APLDet \cite{tang_apldet_2024}, SoftER Teacher \cite{tran2024ledetection} and work of Xiong et al. \cite{xiong2021semi} have attempted similar few-shot settings in the context of SSOD, they rely on pseudo-labeling \cite{lee_2013_pseudolabel} and consistency regularization to identify {\it novel} categories rather than {\it all} classes that appear in a task. For instance, Xiong et al. \cite{xiong2021semi} build upon prior work in meta-learning-based SSOD to improve robustness in a few-shot setting \cite{xiong2021semi}, while APLDet \cite{tang_apldet_2024} uses class-adaptive thresholding to improve pseudo-label quality, achieving SOTA performance under few-shot settings.

\section{Problem Statement}

Training object detectors typically requires large-scale annotated datasets with thousands of labeled instances per class. However, this requirement is impractical in many real-world scenarios, especially in industry applications where only a small number of labeled images per class may be available. In such data-scarce settings, SSOD methods aim to reduce annotation costs by leveraging large sets of unlabeled data.

Existing SSOD models, however, are benchmarked predominantly using fixed percentages of labeled data from standard datasets like MS-COCO. These percentages often correspond to large absolute numbers of object annotations, which is not reflective of true few-shot constraints in terms of images or annotated object instances. Moreover, recent methods that explore SSOD in few-shot regimes generally focus on novel class detection and do not disclose inference-time and resource requirements, hindering their practical applicability.

Our goal is to investigate how state-of-the-art SSOD models—MixPL, Semi-DETR, and Consistent-Teacher—perform under few-shot constraints with a fixed number of randomly sampled labeled images per category. We also analyze inference latency in relation to the model size. 

\section{Methodology}

We propose an empirical benchmarking framework for evaluating SSOD models in realistic few-shot settings, as described in what follows.

\subsection{Few-Shot Sampling Strategy}
Instead of using a fixed percentage of labeled data, we define a fixed number of labeled instances per object category (\textit{k-shot}) and vary $k \in \{1, 5, 10, 20, 50, 100, 150\}$. We apply this to the MS-COCO dataset (with complex multi-object scenes), Pascal VOC and Beetle datasets. We should note that all object instances from the set of $k\times c$ images included in the few-shot setting are used for training. As each image can include several object instances from the same or different categories, especially in the case of MS-COCO and PASCAL VOC datasets, the total number of object instances included in the labeled set is generally much larger than $k\times c$, and exhibits class imbalance.

\subsection{SSOD Approaches}

We select three state-of-the-art SSOD methods—MixPL, Consistent-Teacher, and Semi-DETR—from the previous research \cite{shehzadi2024semi,shehzadi2025step}. These methods span different learning paradigms like augmentation-based, meta-consistent, and transformer-based, offering a broad lens for evaluation under few-shot manner.

\noindent
\textbf{MixPL} \cite{chen_mixed_2023} builds on the Detection Mean Teacher framework \cite{tarvainen_mean_teacher_2017} and introduces two key augmentations - pseudo-Mixup and pseudo-Mosaic - to address challenges in detecting small and tail-class objects. Pseudo-Mixup overlays low-confidence (negative) and high-confidence (positive) pseudo-labels to regularize learning, while pseudo-Mosaic combines four pseudo-labeled samples into a single training instance to enrich supervision. A confidence threshold filters noisy labels, and extensive evaluations across multiple detector backbones confirm its agnostic property. We select MixPL for its strong performance across labeled data regimes and its robustness to noise. 

\noindent
\textbf{Semi-DETR} \cite{zhang_semi-detr_2023} is the first transformer-based SSOD framework, designed to address pseudo-label instability in DETR-style architectures. It introduces Stage-wise Hybrid Matching (SHM) to switch from one-to-many assignment (early stage) to one-to-one matching (late stage), improving pseudo-label quality and training stability. Cross-view Query Consistency (CQC) enforces semantic feature invariance across augmented views without requiring exact object query matching. Additionally, the Cost-based Pseudo Label Mining (CPM) module uses matching costs and Gaussian modeling to filter high-confidence pseudo-boxes. This transformer design provides a strong baseline for evaluating SSOD performance in data-scarce scenarios.

\noindent
 \textbf{Consistent-Teacher} \cite{consistent_teacher} is a meta-consistency-based SSOD framework that addresses the instability of pseudo-labels through three key modules: Adaptive Sample Assignment (ASA) for better anchor matching, 3D Feature Alignment Module (FAM-3D) to reduce spatial misalignment, and a Gaussian Mixture Model (GMM) for dynamic confidence thresholding. Together, these components mitigate overfitting caused by noisy pseudo-boxes and improve the reliability of supervision signals during training. Its design makes it particularly suitable for few-shot regimes where label quality is paramount.

\begin{table*}[t]
  \centering
  \caption{Architecture details and backbones for the SSOD approaches - MixPL, Semi-DETR and Consistent-Teacher}
  \label{backbone}
  \begin{tabular}{l|l|l|l}
  \toprule
  Approach & \shortstack{Detector\\ Architecture} & Backbone & \shortstack{Additional\\ Component} \\ \hline
  MixPL & \shortstack{DINO (DETR-style\\ transformer detector)} & ResNet-50 & N/A \\  \hline
  Semi-DETR & \shortstack{DINO (DETR-style\\ transformer detector)} & ResNet-50 & \shortstack{Semi-supervised wrapper\\ (DinoDetrSSOD)} \\ \hline
  \shortstack{Consistent\\ Teacher} & RetinaNet (single-stage) & ResNet-50 & \shortstack{Feature Pyramid Network (FPN),\\ Mean Teacher framework} \\
  \bottomrule
  \end{tabular}
\vspace{-5mm}
\end{table*}

Table \ref{backbone} provides additional information about the architectures used in the three approaches - MixPL, Semi-DETR and Consistent-Teacher. All three approaches use a pre-trained ResNet-50 network as backbone. MixPL and Semi-DETR have a DETR-style transformer architecture as the detector, specifically, DINO \cite{zhang2022dino}, while Consistent-Teacher employs a lighter-weight RetinaNet \cite{lin2017focal} architecture. This choice of models allows us to gain insights into the use of transformer-based architectures (MixPL/Semi-DETR) versus a lighter-weight architecture (Consistent-Teacher) for SSOD, while also allowing for a comparison of two transformer-based SSOD architectures (MixPL versus Semi-DETR).

\section{Experimental Setup}

\subsection{Model Training}

We evaluate and compare three state-of-the-art SSOD approaches: \textbf{MixPL}, \textbf{Semi-DETR}, and \textbf{Consistent-Teacher}, selected for their complementary architectural designs and prior strong performance in data-limited settings. All models are trained on top of pre-trained ResNet-50 models. The training configurations (batch size, optimizer, learning rate, augmentation) follow the official default settings. We use exactly the same splits/instances for all the models to ensure that any performance changes come only from the model itself, not from different data. For each approach, the evaluation is done using the student model. The models were trained on a system with 4 NVIDIA GPUs A100, 80GB, PCIe and 64 CPU, although not all 4 GPUs were used in each experiment.

\subsection{Datasets}

For the \textbf{MS-COCO dataset} \cite{lin_microsoft_2015}, each $k$ few-shot experiment setting contains $k \times 80$ images. Similarly, for \textbf{Pascal VOC} \cite{everingham2015pascal} and \textbf{Beetle} datasets \cite{wang_detecting_2023}, each $k$ few-shot experiment setting contains $k \times 20$ images and $k \times 7$ images, respectively. Each larger subset, corresponding to a larger $k$, is built on top of the previous one. In terms of object instances, each $k$ few-shot experiment contain a variable number of object instances, due to object co-occurrences in the datasets considered. For example, in the case of MS-COCO, the total number of instances is 324 for the 1-shot experiment, 1,512 for the 5-shot experiment, etc. Common categories such as \textit{person}, \textit{chair}, and \textit{cup} dominate due to their over-representation in the dataset. This structure highlights how the annotations capture greater diversity and better simulate noisy real-world distributions. 

The exact counts of images and object instances, for each $k$-shot experiment are shown in Table \ref{table_few_shots}, specifically, the columns labeled \#Total Images and \#Annotated Instances, respectively. All models are evaluated on the MS-COCO, Pascal VOC and Beetle respective test subsets. This allows for relative comparisons with fully supervised results on the three datasets, as well as prior SSOD results available for MS-COCO and Pascal VOC datasets. Statistics about the train/validation/test splits of the three datasets and percentages of data used in prior works for MS-COCO are shown in Table \ref{table_all_dataset_baselines}. 

A comparison between our few-shot subsets (Table \ref{table_few_shots}) and subsets used in prior SSOD works on MS-COCO (top panel in Table \ref{table_all_dataset_baselines}), shows that the number of images in the COCO 10\% subset (specifically, 11,828) is comparable to the total number of images in our 150-shot per class experiment (which is 12,000). However, in terms of instances, COCO 10\% has almost twice as many object instances as compared to the 150-shot per class experiment, while the number of instances in COCO 5\% (42,962) is comparable with the number of instances in the 150-shot experiment (45,984) and allows for a more direct comparison.

\begin{table}[h]
\centering
\caption{Few shot information on MS-COCO, Pascal VOC and Beetle datasets: For each $k$-shot, the total number of images (\#Total Images) and the number of annotated instances (\#Annot. Inst.) used for each experiment are shown.}
\label{table_few_shots}
\begin{tabular}{ r| r|r | r|r | r|r }

\toprule

& \multicolumn{2}{c|}{\textbf{MS-COCO}} & \multicolumn{2}{c|}{\textbf{Pascal VOC}} & \multicolumn{2}{c}{\textbf{Beetle}} \\

\midrule

\textbf{\shortstack{\#Shots/ \\ class}} & \textbf{\shortstack{\#Total\\Images}} & \textbf{\shortstack{\#Annot. \\Inst.}} & \textbf{\shortstack{\#Total\\Images}} & \textbf{\shortstack{\#Annot. \\Inst.}} & \textbf{\shortstack{\#Total\\Images}} & \textbf{\shortstack{\#Annot. \\Inst.}} \\

\midrule

  1 &     80 &    324 &    20 &     73 &     7 &     8 \\
  5 &    400 &  1,512 &   100 &    311 &    35 &    38 \\
 10 &    800 &  3,057 &   200 &    722 &    70 &    76 \\
 20 &  1,600 &  6,184 &   400 &  1,517 &   140 &   147 \\
 50 &  4,000 & 15,423 & 1,000 &  3,503 &   350 &   716 \\
100 &  8,000 & 30,852 & 2,000 &  6,903 &   700 &   746 \\
150 & 12,000 & 45,984 & 3,000 & 10,104 & 1,050 & 1,112 \\

\bottomrule
\end{tabular}
\vspace{-5mm}
\end{table}


\begin{table}[h]
\centering
\caption{Data statistics for the Train/Val/Test subsets of the MS-COCO, Pascal VOC and Beetle datasets, respectively. Statistics for the percentage of data used in prior SSOD works are also shown for the MS-COCO dataset. The statistics are shown both in terms of number of total images in a subset, as well as the total number of annotated instances in the subset.}
\label{table_all_dataset_baselines}

\begin{tabular}{l|r|r}

\toprule

Dataset & \#Total Images & \#Annotated Instances \\

\midrule

\multicolumn{3}{c}{\textit{MS-COCO}} \\
\hline
\hline

COCO 1\%   &   1,182 &   8,762 \\
COCO 2\%   &   2,366 &  16,751 \\
COCO 5\%   &   5,914 &  42,962 \\
COCO 10\%  &  11,828 &  86,066 \\
COCO 100\% & 118,287 & 860,001 \\
Val        &   5,000 &  36,781\\
Test       &  40,670 & N/A \\

\midrule

\multicolumn{3}{c}{\textit{Pascal VOC}} \\
\hline
\hline

VOC07 train/val &  5,011 & 12,608 \\
VOC12 train/val & 11,540 & 31,561 \\
VOC07 test        &  4,952 & 12,032 \\

\midrule

\multicolumn{3}{c}{\textit{Beetle}} \\
\hline
\hline

Train & 3,053 & 3,232 \\
Val & 1,113 & 1,176 \\ 
Test & 699 & 734 \\

\bottomrule
\end{tabular}
\vspace{-5mm}
\end{table}

\begin{table}[h]
\centering
\caption{{\bf Fully Supervised and SSOD baselines from prior works.} SSOD baseline results for the MS-COCO dataset are based on different percentages of labeled data, specifically 1\%, 2\%, 5\%, 10\% and 100\%. SSOD baseline results for Pascal VOC use a subset of the VOC07 set as labeled data, VOC12 as unlabeled data and a subset of VOC07 as test data. No prior SSOD results are available for the Beetle dataset. Supervised baseline results are also reported for all three datasets. Performance is reported in terms of mAP[0.50:0.95] (mAP).}
\label{table_all_baselines}
\scriptsize 
\renewcommand{\arraystretch}{0.85} 
\resizebox{\textwidth}{!}{ 

\begin{tabular}{c|l|r|r|r|r|r}
\toprule

\multicolumn{7}{c}{\textit{MS-COCO}} \\

\hline
\midrule

  & \% of MS-COCO &  1\% &  2\% &  5\% &  10\% &  100\% \\ \midrule
\multirow{5}{*}{SSOD}
  & Consistent-Teacher \cite{consistent_teacher} & 25.30 & 30.40 & 36.10 & 40.00 & 47.70 \\
  & Semi-DETR DINO \cite{zhang_semi-detr_2023} & 30.50 $\pm$ 0.30 & - & 40.10 $\pm$ 0.15 & 43.50 $\pm$ 0.10 & 50.40 \\
  & MixPL DINO \cite{chen_mixed_2023} & {\bf 31.70} & \textbf{34.70} & 40.10 & 44.60 & \textbf{55.20} \\ 
  & Sparse Semi-DETR \cite{shehzadi2024sparse} & 30.90 $\pm$ 0.23 & - & 40.80 $\pm$ 0.12 & 44.3 $\pm$ 0.01 & 51.30 \\
  & STEP-DETR \cite{shehzadi2025step} & \textbf{31.70 $\pm$ 0.30} & - & \textbf{41.10 $\pm$ 0.11} & \textbf{45.40 $\pm$ 0.10} & 52.10 \\

\midrule

\multirow{3}{*}{\shortstack{Fully\\ Supervised OD}}
    & YOLOv13-X \cite{lei2025yolov13}   &  &  &  &  & 54.80 \\
    & Faster RCNN \cite{Shaoqing_faster_rcnn} &  &  &  &  & 21.90 \\
    & DETR  \cite{detr_nicolas}      &  &  &  &  & 44.90 \\
  
\midrule
\multicolumn{7}{c}{\textit{Pascal VOC}} \\

\hline
\midrule
  &            VOC07 train/val         &  & {\color{white}mAP.50} &  & &  100\%\\ \midrule

\multirow{7}{*}{SSOD} 
  & Consistent-Teacher \cite{consistent_teacher}   &  & {\color{white}81.00} &  && 59.00 \\ 
  & Semi-DETR Def-DETR \cite{zhang_semi-detr_2023} &  & {\color{white}83.50} &  && 57.20 \\ 
  & Semi-DETR DINO \cite{zhang_semi-detr_2023}     &  & {\color{white}86.10} &  && 65.20 \\ 
  & MixPL Faster R-CNN \cite{chen_mixed_2023}      &  & {\color{white}85.80} &  && 56.10 \\ 
  & MixPL FCOS \cite{chen_mixed_2023}              &  & {\color{white}84.70} &  && 59.00 \\

  & Sparse Semi-DETR \cite{shehzadi2024sparse}     &  & {\color{white}86.30} &  && 65.51 \\ 
  & STEP-DETR \cite{shehzadi2025step}              &  & {\color{white}{\bf 86.85}} &  && {\bf 65.87} \\ 

\midrule

\multirow{3}{*}{\shortstack{Fully\\ Supervised OD}}
    &&&&&& \\
    & Faster RCNN \cite{Shaoqing_faster_rcnn} &  &  &   & & 59.90\\
    &&&&&& \\

\midrule

\multicolumn{7}{c}{\textit{Beetle}} \\

\hline
\midrule
  &                         & {\color{white}mAP} & {\color{white}mAP.50} & {\color{white}mAP.75} &  & 100\% \\ 
\midrule

\multirow{3}{*}{\shortstack{Fully \\Supervised OD}}
    & Faster-R101-GIoU \cite{wang_detecting_2023}&  & {\color{white}93.7} & {\color{white}75.60 }&  & 65.60 \\
    & YOLOv5x \cite{wang_detecting_2023}& & {\color{white}95.9} & {\color{white}85.6} &  & 73.80  \\
    & YOLOv7 \cite{wang_detecting_2023}& & {\color{white}97.3} & {\color{white}86.2} &  & 74.60  \\

\bottomrule
\end{tabular}
}
\vspace{-5mm}
\end{table}

\subsection{Evaluation Metrics}

Experimental results are reported in terms of standard object detection metrics, such as mean Average Precision (mAP) averaged over all thresholds from [0.5:0.95] with an increment of 0.5. We also report approximate inference time (ms/image), as well as average model size (in MB). This systematic evaluation highlights trade-offs between detection quality and deployment efficiency.

Each model is tested across the three datasets, enabling us to assess how they respond to data scarcity. Our setup allows for robust comparison of:
\begin{itemize}
\item The model's inference latency for models trained in a SSOD low-data regime.
\item The generalization capacity as the number of annotations $k$ increases.
\end{itemize}

\section{Results \& Discussion}

Fully supervised and previously reported SSOD baseline results are summarized in Table~\ref{table_all_baselines} for MS-COCO, Pascal VOC, and the Beetle dataset. The results of our few-shot SSOD experiments, obtained under explicitly controlled $k$-shot-per-class supervision, are reported in Table~\ref{table_all_results_merged}. In contrast to prior SSOD studies that define supervision using coarse dataset percentages (e.g., COCO 1\%, 5\%, or 10\%), our evaluation protocol enforces balanced per-class coverage, ensuring that each object category appears in at least $k$ labeled images. This setting more closely reflects realistic data-scarce scenarios encountered in practice and enables a more fine-grained analysis of label efficiency.

\begin{figure}[h]
\centering
\includegraphics[width=0.9\linewidth]{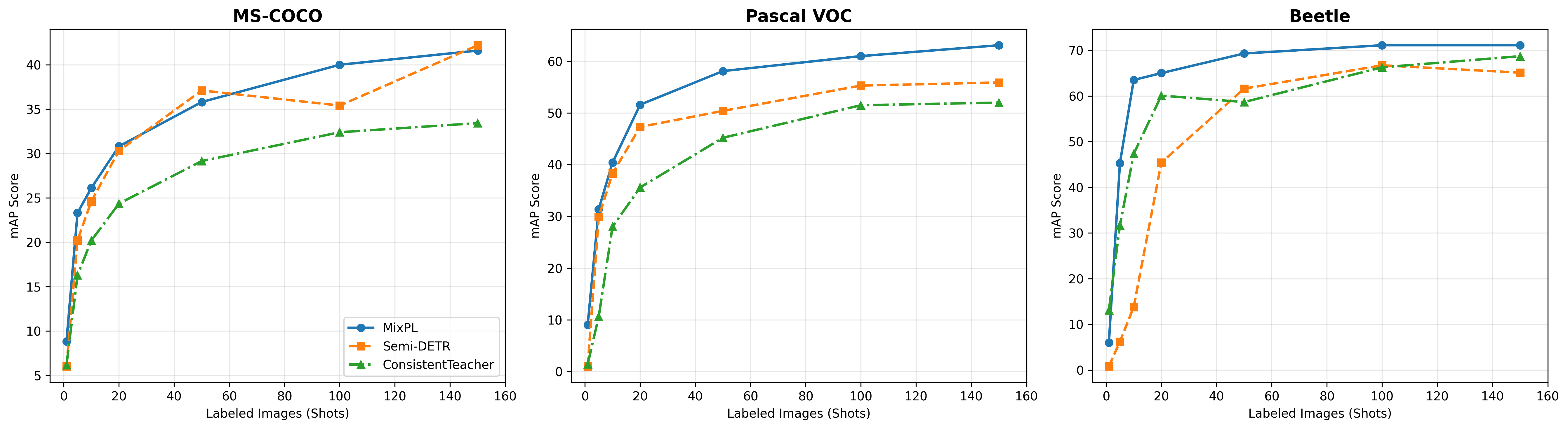}
\caption{Performance comparisons with the number of $k$-shots across MixPL, Semi-DETR, and Consistent-Teacher on MS-COCO, {Pascal VOC and Beetle datasets}.}
\label{fig:strict_results}
\end{figure}

Following prior SSOD works, the results are obtained with the student model, which is trained on both labeled and pseudo-labeled data. In addition to mAP performance, we also report in Table ~\ref{table_all_results_merged}, approximate inference time and model size for each model on each dataset. Fig. \ref{fig:strict_results} shows the variation of performance with the number of shots/images used for training for visual analysis.

\begin{table}[h]
\centering
\caption{SSOD results of MixPL, Semi-DETR and Consistent-Teacher models for different $k$-shots per class for the three datasets used (MS-COCO, Pascal VOC, and Beetle). Results are reported in terms of {\bf mAP} (mAP[0.50:0.95]). In addition, approximate inference {\bf time} (milliseconds per image, ms/img) and model {\bf size} (MB) are shown for the three models and the three datasets. }
\label{table_all_results_merged}

\begin{tabular}{ c| c| r c c| r c c| r c c }
\toprule
\textbf{Model} &
\textbf{\shortstack{\#Shots/\\ class}} &
\multicolumn{3}{c|}{\textbf{MS-COCO}} &
\multicolumn{3}{c|}{\textbf{Pascal VOC}} &
\multicolumn{3}{c}{\textbf{Beetle}} \\

\cmidrule(lr){3-5}\cmidrule(lr){6-8}\cmidrule(lr){9-11}
& &
\textbf{mAP} &
\textbf{\shortstack{Time}} &
\textbf{\shortstack{Size}} &
\textbf{mAP} &
\textbf{\shortstack{Time}} &
\textbf{\shortstack{Size}} &
\textbf{mAP} &
\textbf{\shortstack{Time}} &
\textbf{\shortstack{Size}} \\

\midrule

\multirow{7}{*}{\textbf{MixPL}}
 &   1 &  8.80 & \multirow{7}{*}{37.20} & \multirow{7}{*}{920}
        &  9.00 & \multirow{7}{*}{36.02} & \multirow{7}{*}{830}
        &  6.00 & \multirow{7}{*}{34.97} & \multirow{7}{*}{919} \\
 &   5 & 23.30 & & & 31.40 & & & 45.30 & & \\
 &  10 & 26.10 & & & 40.40 & & & 63.50 & & \\
 &  20 & 30.80 & & & 51.60 & & & 65.00 & & \\
 &  50 & 35.80 & & & 58.10 & & & 69.30 & & \\
 & 100 & 40.00 & & & 61.00 & & & \textbf{71.10} & & \\
 & 150 & \textbf{41.60} & & & \textbf{63.10} & & & \textbf{71.10} & & \\

\midrule

\multirow{7}{*}{\textbf{\shortstack{Semi-\\DETR}}}
 &   1 &  6.00 & \multirow{7}{*}{43.00} & \multirow{7}{*}{885}
        &  1.00 & \multirow{7}{*}{42.40} & \multirow{7}{*}{884}
        &  0.80 & \multirow{7}{*}{40.00} & \multirow{7}{*}{884} \\
 &   5 & 20.20 & & & 29.90 & & &  6.20 & & \\
 &  10 & 24.60 & & & 38.30 & & & 13.80 & & \\
 &  20 & 30.30 & & & 47.30 & & & 45.40 & & \\
 &  50 & 37.10 & & & 50.40 & & & 61.60 & & \\
 & 100 & 35.40 & & & 55.30 & & & \textbf{66.70} & & \\
 & 150 & \textbf{42.20} & & & \textbf{55.90} & & & 65.10 & & \\

\midrule

\multirow{7}{*}{\textbf{\shortstack{Consistent-\\Teacher}}}
 &   1 &  6.13 & \multirow{7}{*}{9.80} & \multirow{7}{*}{372}
        &  1.40 & \multirow{7}{*}{10.40} & \multirow{7}{*}{387.5}
        & 13.08 & \multirow{7}{*}{15.30} & \multirow{7}{*}{370} \\
 &   5 & 16.27 & & & 10.60 & & & 31.65 & & \\
 &  10 & 20.18 & & & 28.00 & & & 47.33 & & \\
 &  20 & 24.34 & & & 35.60 & & & 60.07 & & \\
 &  50 & 29.14 & & & 45.20 & & & 58.66 & & \\
 & 100 & 32.39 & & & 51.50 & & & 66.27 & & \\
 & 150 & \textbf{33.42} & & & \textbf{52.00} & & & \textbf{68.70} & & \\

\bottomrule
\end{tabular}
\vspace{-5mm}
\end{table}

\subsection{Performance Analysis Across Datasets}

\paragraph{MS-COCO.}
On the MS-COCO dataset, all three SSOD approaches exhibit consistent performance improvements as the number of labeled images per class increases. Among the evaluated methods, MixPL achieves the strongest overall performance across all $k$-shot regimes, reaching \textbf{41.6 mAP} at 150 shots per class. Notably, this performance exceeds the COCO 5\% SSOD baseline reported in Table~\ref{table_all_baselines} (40.1 mAP), despite using fewer total labeled images. Importantly, the 150-shot setting contains a comparable number of annotated object instances (approximately 46k) to COCO 5\% (approximately 43k), indicating that \emph{structured per-class supervision yields stronger performance than percentage-based sampling at similar annotation cost}.

While performance at 150 shots remains below the COCO 10\% baseline (44.6 mAP for MixPL), the gap is modest given the substantially smaller labeled image pool and the stricter supervision constraints. Semi-DETR follows a similar trend, reaching \textbf{42.2 mAP} at 150 shots, demonstrating competitive performance relative to several COCO 5\%--10\% SSOD baselines. Consistent-Teacher, while lagging in absolute accuracy, still shows steady gains with increasing $k$, confirming its robustness in low-data regimes.

Across all models, the most pronounced improvements occur in the low-shot range (1--50 shots), with diminishing returns observed beyond 100 shots. This behavior suggests that, for complex multi-object datasets such as MS-COCO, early gains are driven primarily by achieving minimal class coverage, while later improvements require disproportionately more labeled data.

\paragraph{Pascal VOC.}
On Pascal VOC, performance improves rapidly with relatively few labeled images per class, reflecting the dataset's lower visual complexity and smaller number of categories. MixPL reaches \textbf{63.1 mAP} at 150 shots per class, closely matching prior SSOD results obtained using significantly larger labeled subsets (Table~\ref{table_all_baselines}). Semi-DETR and Consistent-Teacher exhibit similar saturation behavior, with gains tapering beyond 100 shots.

These results indicate that, for moderately complex datasets, SSOD models can achieve near-saturated performance with far fewer labeled images than typically assumed in percentage-based supervision protocols. The early saturation further highlights the efficiency of balanced per-class annotation strategies.

\paragraph{Beetle Dataset.}
The Beetle dataset exhibits the strongest few-shot learning behavior across all models. With only seven object categories and visually homogeneous scenes, MixPL achieves \textbf{71.1 mAP} by 100 shots per class, after which performance plateaus. Additional labeled data did not yield further improvements, suggesting that the model reaches its representational capacity for this task early.

Consistent-Teacher demonstrates the strongest 1-shot performance on this dataset, indicating that lighter-weight, anchor-based architectures may be advantageous in extremely low-data, low-diversity settings. Semi-DETR shows delayed but substantial gains, particularly between 10 and 20 shots, underscoring the sensitivity of transformer-based SSOD methods to minimal supervision in specialized domains.

\subsection{Key Observations and Implications}

When interpreted jointly, Tables~\ref{table_all_baselines} and~\ref{table_all_results_merged} reveal several important insights. First, enforcing more balanced per-class supervision enables SSOD models to match or exceed traditional percentage-based baselines at comparable annotation cost. Second, the marginal benefit of additional labeled data depends strongly on dataset complexity: simpler datasets saturate quickly, while complex datasets such as MS-COCO continue to benefit from increased supervision. Finally, although transformer-based methods achieve superior peak accuracy, lighter-weight architectures remain competitive in extremely low-shot regimes and offer favorable trade-offs for resource-constrained deployments.


Overall, these findings demonstrate that $k$-shot per-class evaluation provides a more realistic and informative lens for assessing SSOD performance in data-scarce scenarios, offering practical guidance for annotation strategy design in real-world applications.

\subsection{Inference Time}

Across all datasets, Consistent-Teacher consistently achieves the lowest inference latency, processing images in approximately 9--15 ms per image, while MixPL and Semi-DETR require approximately 35--43 ms per image. These differences align with architectural choices: Consistent-Teacher relies on a single-stage CNN-based detector, whereas MixPL and Semi-DETR employ transformer-based architectures with substantially higher computational overhead. Importantly, inference time remains stable across $k$-shot settings, confirming that latency is dominated by model architecture rather than training data volume. When considered jointly with the accuracy results in Table~\ref{table_all_results_merged}, these findings reveal a clear performance--latency trade-off: transformer-based SSOD models deliver higher peak accuracy, particularly in mid-to-high data regimes, at the cost of approximately 3--4$\times$ higher inference latency compared to Consistent-Teacher.

\subsection{Model Footprint}

When considered alongside the inference time and accuracy results (Tables~\ref{table_all_results_merged} and~\ref{table_all_baselines}), model footprint further clarifies the trade-offs among SSOD approaches. The larger memory footprint of MixPL and Semi-DETR reflects the cost of transformer-based architectures, which consistently achieve higher peak accuracy in mid-to-high data regimes. In contrast, Consistent-Teacher maintains a substantially smaller size, enabling faster inference and lower memory requirements, but with reduced performance. These results indicate that model size, latency, and accuracy form a coupled design space, and that architecture choice should be guided by deployment constraints as much as by accuracy targets.

\subsection{Research Questions and Practical Recommendations}

The above analysis allows us to answer the questions raised in the Introduction.

{\it RQ1: What is the best SSOD when the number $k$ of labeled images per category varies between 1 and 150?} Overall, the MixPL model has superior performance compared to the other models across different k-shot settings. The Semi-DETR model is closely behind, and sometimes slightly better than the MixPL model, while the Consistent-Teacher generally has lower performance. This is not surprising given that the base model for MixPL and Semi-DETR is DINO - a transformer-based model, while the base model for Consistent-Teacher is a CNN model. This is also reflected in the size of the models. 

{\it RQ2: What are the trade-offs between low-regime data training and overall object detection performance?} Analyzing the variations in mAP across increasing numbers of $k$-shots, as illustrated in Fig. \ref{fig:strict_results} and detailed in {Table \ref{table_all_results_merged}}, clearly demonstrates a positive correlation between labeled data availability and detection performance, although there are some cases where performance decreases with the number of labeled instances possibly due to noise in labeled or pseudo-labeled data. Performance gains are most significant when transitioning from extremely low data regimes (1-shot, 5-shot) towards intermediate levels (50-shot, 100-shot), with incremental gains beyond 100-shot settings. Notably, in low-data regimes, Consistent-Teacher achieves accuracy closer to MixPL and Semi-DETR, whereas the performance gap widens significantly with more labeled data. Thus, Consistent-Teacher provides a reasonable alternative in extremely limited labeled data scenarios, but transformer-based architectures yield notably higher returns in detection accuracy as labeled data availability increases.

{\it RQ3: What are the trade-offs between performance, model size and latency?} The experimental results reveal clear trade-offs between detection accuracy, model size, and latency (inference time) as inferred from Table ~\ref{table_all_results_merged}. MixPL and Semi-DETR deliver superior accuracy, particularly in mid-to-high data regimes, but incur greater computational overhead (due to larger size) and higher latency during inference (MixPL: 37.2 ms/image, Semi-DETR: 43 ms/image). This latency could be prohibitive for real-time or edge-device deployments. In contrast, Consistent-Teacher exhibits significantly lower latency (9.8 ms/image), making it highly attractive for applications with real-time constraints or limited computational resources, despite its comparatively lower accuracy. Therefore, selecting the appropriate model depends critically on practical constraints: for high-accuracy demands with sufficient computational resources, MixPL or Semi-DETR are optimal choices; for latency-sensitive and resource-constrained environments, Consistent-Teacher provides a highly favorable trade-off.

\subsection{Practical Recommendation}
The results of our study suggest that the transformer models, and especially MixPL, achieve competitive results with a relatively small number of labeled images, which makes them ideal in a low-data regime. However, the size of the transformer models is significantly larger as compared to the size of the Consistent-Teacher, and thus the resources needed for training these models are also more expensive. Additionally, the models incur higher latency. In very low-data and low-resource regimes, the Consistent-Teacher may be a good alternative. 
While our study was specifically focused on the comparison of three SSOD frameworks with transformer or CNN-based backbones, we anticipate that the observed trade-offs between accuracy, model size and latency will generalize to other newer SSOD frameworks and architectures, which will likely shift the performance-latency Pareto frontier upwards. However, the relative trend in which  transformer-based SSOD methods favor accuracy, while CNN-based methods favor speed is expected to persist.

\section{Conclusions, Limitations and Future Work}

A series of experiments were conducted for a systematic evaluation of three state-of-the-art SSOD models (MixPL, Semi-DETR and Consistent-Teacher) on MS-COCO, Pascal VOC and Beetle datasets. Our study explored the models in a practically relevant data regime through a few-shot setting where only a small number of annotated examples are available. Our results reveal that while these models demonstrate some capacity to generalize in low-data scenarios, their performance degrades significantly as the number of labeled instances decreases. 

{Our study is not limited to the exploration of performance but also focuses on practical considerations such as inference time and model size. Overall, our findings provide insights into the applicability and limitations of current SSOD models in resource-constrained settings, by using several datasets that exhibit smaller or larger number of classes and class imbalance. However, scaling our framework to datasets with significantly larger category counts (e.g., LVIS \cite{gupta2019lvis}) or long-tail distributions may present additional challenges.}

Future work will explore the behavior of the SSOD models and Large Language Models (LLMs)/Vision Language Models (VLMs) for other types of datasets and distributions prevalent in various application domains. For example, it is of interest to explore other object detection datasets with a small number of objects and a small number of object instances in an image, or datasets with a small number of objects but a large number of object instances in an image, as well as datasets with larger number of objects and long-tail distributions.

\section*{Acknowledgments}

The authors would like to thank Peak Technologies for providing access to computational resources that supported this research. 

\bibliographystyle{splncs04}
\bibliography{references}

\end{document}